\documentclass[letterpaper, 10 pt, conference]{ieeeconf}  

\IEEEoverridecommandlockouts                              

\usepackage{amsmath} 
\usepackage{amssymb}  

\usepackage{graphicx}
\usepackage{subcaption}

\usepackage{tabularx}
\usepackage{booktabs}

\usepackage{rotating}

\usepackage{multirow,array}
\usepackage{gensymb}
\usepackage{adjustbox}

\usepackage{makecell}

\usepackage{xcolor}
\usepackage{xspace}

\usepackage{hyperref}

\usepackage[bottom]{footmisc}

\DeclareMathOperator*{\argmin}{arg\,min}

\DeclareMathOperator*{\SE3}{\mathbb{SE}(3)}

\def\ie{\emph{i.e.}\xspace}

\newcommand\bblue[1]{\textcolor{blue}{\textbf{#1}}}
\newcommand\blue[1]{\textcolor{blue}{#1}}
\newcommand\itbf[1]{\textit{\textbf{#1}}}

\usepackage{caption}
\def\ov2slam{OV$^{2}$SLAM}
\def\smov2slam{ov$^{2}$slam}
\def\fov2slam{OV$^{2}$SLAM-Fast}
\def\fsmov2slam{ov$^{2}$slam-fast}

\title{\LARGE \bf OV$^{2}$SLAM : A Fully Online and Versatile Visual SLAM for Real-Time Applications}

\author{Maxime Ferrera$^{1,2,\dagger}$, Alexandre Eudes$^{1}$, Julien Moras$^{1}$, Martial Sanfourche$^{1}$ and Guy Le Besnerais$^{1}$
\thanks{$^{*}$This work was supported by the ANR/DGA project MALIN.
\linebreak
$^{1}$ DTIS, ONERA, Université Paris-Saclay, F-91123, Palaiseau, France. Contact emails : {\tt\footnotesize \{first.last@onera.fr\}} %
\linebreak
$^{2}$ IFREMER, Ctr. Méditerranée, Underwater System Unit, CS20330, F-83507, La-Seyne-Sur-Mer, France. {\tt\footnotesize \{first.last@ifremer.fr\}}%
\linebreak
$^{\dagger}$ work done while at ONERA.
}}

\begin{document}

\maketitle
\thispagestyle{empty}
\pagestyle{empty}

\begin{abstract}

Many applications of Visual SLAM, such as augmented reality, virtual reality, robotics or autonomous driving, require versatile, robust and precise solutions, most often with real-time capability. In this work, we describe  \ov2slam, a fully online algorithm, handling both monocular and stereo camera setups, various map scales and frame-rates ranging from a few Hertz up to several hundreds. It combines numerous recent contributions in visual localization within an efficient multi-threaded architecture. Extensive comparisons with competing algorithms shows the state-of-the-art accuracy and real-time performance of the resulting algorithm.  For the benefit of the community, we release the source code: \url{https://github.com/ov2slam/ov2slam}.

\end{abstract}

\section{Introduction}

\begin{figure}[htbp]
\begin{center}
    \includegraphics[width=0.95\linewidth]{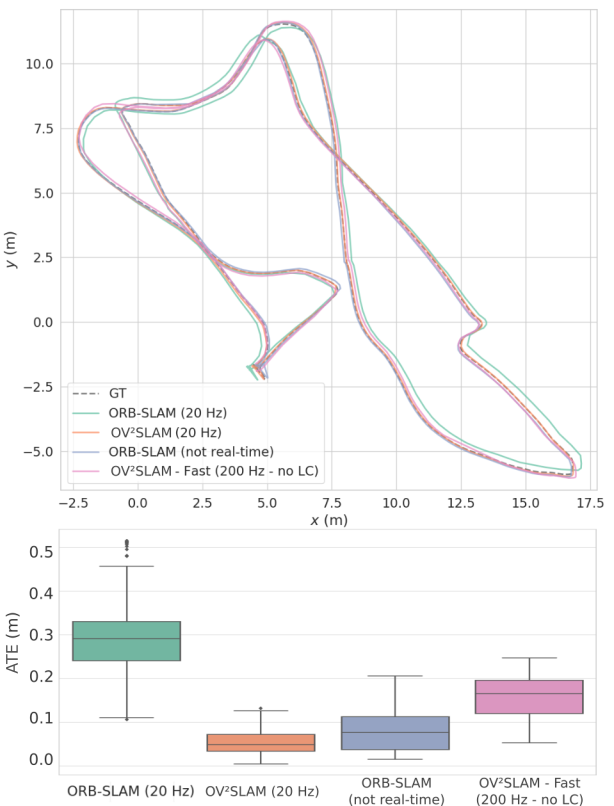}
\end{center}
\caption{Stereo VSLAM results on MH05 trajectory of EUROC dataset. When RT is enforced the mean ATE of ORB-SLAM is multiplied by 3 and the dispersion of its results increases significantly. While running at 20Hz, \ov2slam outperforms both ORB-SLAM results. Besides, a fast version compatible with operation at 200Hz still leads to a high accuracy.}
\vspace{-5mm}
\label{fig:tradeoff}
\end{figure}

Nowadays Visual SLAM (VSLAM) is getting more and more mature.
Yet, state-of-the-art methods still struggle to achieve simultaneously accuracy, robustness and real-time (RT) capability.
In the context of VSLAM, the RT constraint is related to the camera's frame-rate and, in practice, the problem comes from frame losses related to peaks in the processing time. Even if an algorithm processes images faster or at the camera's frame-rate \textit{in average}, those peaks implies that information is lost in ``RT forced conditions" where, at each time, the most recently received image is processed. As illustrated in Fig.~\ref{fig:tradeoff} for trajectory MH05 of EUROC dataset~\cite{burri2016euroc}, when tested in such conditions, ORB-SLAM \cite{mur2017orb}
sees its accuracy decrease significantly with respect to ``not RT" conditions, where successive frames are all processed whatever the processing delay of a particular frame. 

In this paper, we describe OV$^{2}$SLAM, a fully online VSLAM algorithm able to process stereo or monocular streams, which aims at closing the gap between RT capability, accuracy and robustness. As shown in Fig~\ref{fig:tradeoff}, \ov2slam outperforms both ORB-SLAM versions while running at 20Hz. Besides, a fast configuration of \ov2slam running at 200Hz remains highly accurate on this dataset.

It is well-known since PTAM \cite{klein2007parallel} that multi-threading is the key to efficient VSLAM implementation. We push this principle further by proposing a four-threads architecture (front-end, mapping, state optimization and loop closing). While most of the operations within each thread stem from previously published methods, our contribution is mainly in a very careful management of these operations so as to minimize drift and save runtime. Besides, the loop closing (LC) thread is based on the online Bag-of-Words (BoW) algorithm iBoW-LCD \cite{Garcia-Fidalgo2018ibow-lcd}. The interest of iBoW-LCD, which, to the best of our knowledge, has not been included in a VSLAM implementation before, is that the vocabulary tree employed for loop detection is built incrementally, making it always suited to the current environment.

\ov2slam has been compared with several competing algorithms on various publicy available datasets, showing  state-of-the-art accuracy while being fully RT.  We further see \ov2slam as an open research platform, hence we release our code to the benefit of the community.

\section{Related Work}\label{sec:soa}

VSLAM has been extensively studied in the past decades, mainly because of the affordability of camera's systems but also because of the great amount of information provided by imaging systems.
PTAM \cite{klein2007parallel} has been one of the first monocular VSLAM algorithm able to employ Bundle Adjustment (BA) while ensuring RT capability on CPU by cleverly decomposing the VSLAM motion and structure problems, allowing to employ a multi-threaded architecture. 
At the time of writing, ORB-SLAM \cite{mur2017orb} is the state-of-the-art VSLAM. We must also consider visual odometry (VO) methods, such as SVO \cite{forster2016svo} and DSO \cite{engel2017direct}, which do not perform LC. All these algorithms run on CPU and have been publicly released, greatly impacting the robotic and computer vision fields.

The aforementioned algorithms can be divided in two categories: feature-based and direct.  Direct approaches jointly estimate the tracking of pixels and the pose of the camera, usually by computing an image alignment through the minimization of a photometric error.  On the other hand, feature-based methods (also called undirect methods) separate the tracking and the pose estimation parts, solving them sequentially and relying on a geometric error minimization for pose estimation.  We refer the interested readers to \cite{engel2017direct} for more details on the difference between both approaches.

DSO falls in the direct VO category.  It is a monocular method based on the tracking of a sparse set of pixels and carefully taking into account the image formation parameters such as gain and lens vignetting.  While the proposed \ov2slam also relies on a photometric tracking, it follows an undirect approach, still more robust than direct ones on most of the existing datasets.  Indeed, while DSO shows impressive results when used with photometrically calibrated cameras, its accuracy drops quickly when used with regular cameras not providing these information \cite{yang2018challenges}.  
SVO is a hybrid VO method, initializing its estimated pose in a direct way but then minimizing a re-projection error to refine the estimated pose.  It handles both monocular and stereo setups and focuses on speed but at the expense of robustness, failing quite often in difficult sequences \cite{forster2016svo}.  

Finally, ORB-SLAM is probably the most complete VSLAM open-sourced algorithm at the time of writing.  
It proposed a feature-based method suited for either monocular, stereo or RGB-D camera setups.  It relies on the extraction of ORB \cite{rublee2011orb} features and use them for all the algorithm tasks.  ORB-SLAM is one of the most accurate VSLAM method up-to-date.  Its impressive accuracy comes from its very Structure-from-Motion (SfM) like pipeline, limiting at most the triangulation of new map points by heavily trying to match extracted features to already existing map points.  This way, it highly reduces the amount of noise included in the 3D map and relies on the induced covisibility constraints between these map points and their related keyframes for optimal optimization through BA.  Furthermore, its LC capability combined with its local map tracking allows to both close small and large loops, keeping the drift extremely low whenever re-visiting an already mapped area.  Yet, this accuracy comes at the price of potentially high run-time requirement, leading to wide drops in terms of accuracy when enforcing RT (see Sec.\ref{sec:expe} and \cite{forster2016svo}).  A common issue with feature-based methods is their need for extracting features in every acquired image.  In \ov2slam, we heavily reduce the computational load by limiting the extraction of features to keyframes and track them in subsequent frames by minimization of a photometric error.  Yet, in opposition to pure direct methods, we use the extracted descriptors for the means of local map tracking such as in ORB-SLAM but only performing this step for keyframes.  This way we manage to close small loops by re-tracking lost or temporarily occluded map points without impacting the front-end speed but gaining in robustness and accuracy.

Furthermore, to the best of our knowledge, all the proposed VSLAM algorithms that integrate a LC feature rely on offline BoW, pre-trained on a given database.  In opposition, for the first time we propose to integrate an online BoW approach that incrementally builds its vocabulary tree from the provided descriptors, making it always suited to the current environment and less impacted by biases related to an off-line training database~\cite{angeli2008fast, nicosevici2012automatic}.  Thus, \ov2slam benefits from the state-of-the-art performance of iBoW-LCD \cite{Garcia-Fidalgo2018ibow-lcd}, making it highly efficient in very diverse environments.

\section{Architecture, assumptions, notations}\label{sec:archi}

\begin{figure}[!t]
\begin{center}
    \includegraphics[width=\columnwidth]{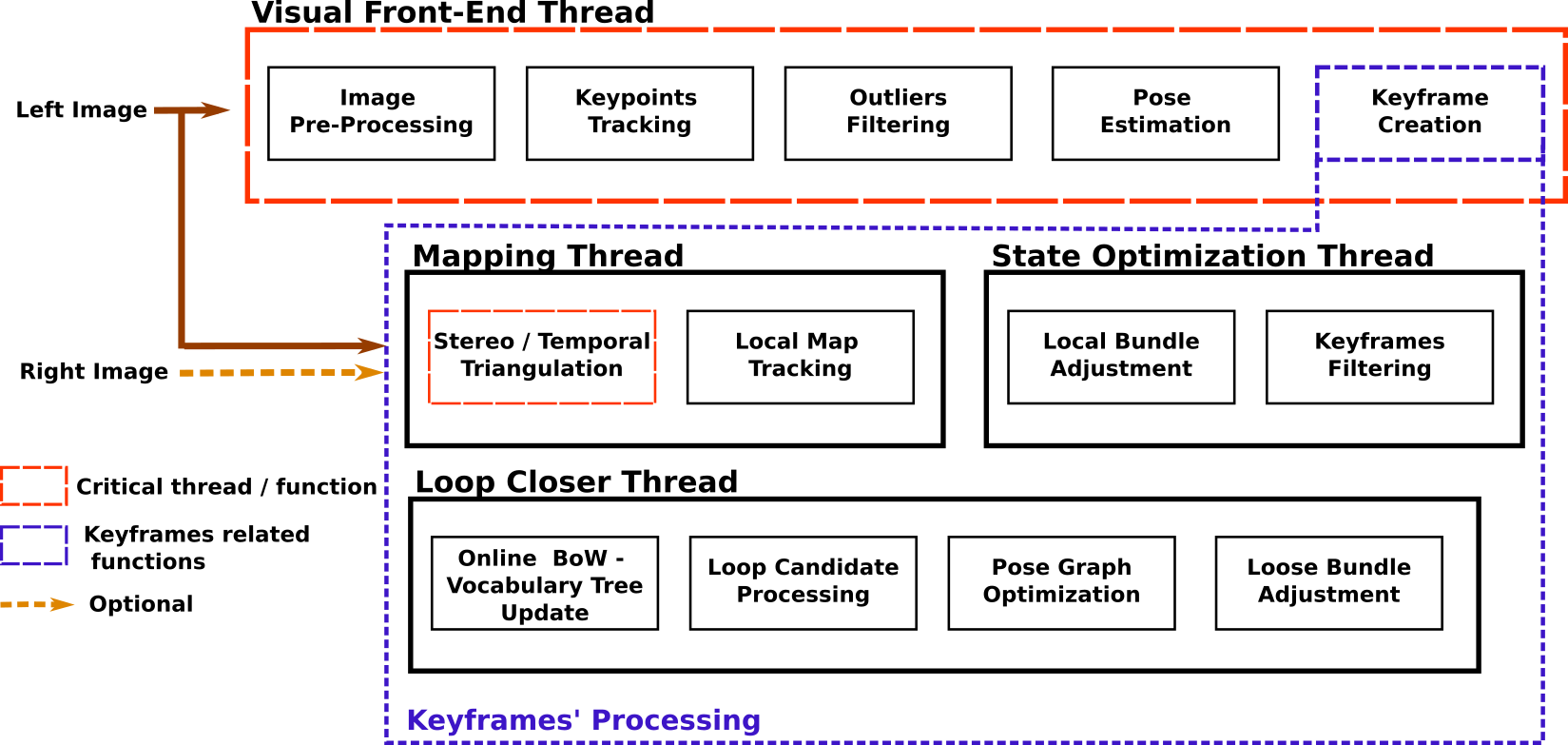}
\end{center}
\caption{Architecture of \ov2slam.  Critical operations in terms of RT processing are highlighted in red.  Functions related to the processing of keyframes are grouped inside the magenta frame: while not critical, the fastest they run, the better we ensure accuracy.}
\vspace{-5mm}
\label{fig:archi}
\end{figure}

\textbf{Architecture.}  The architecture of OV$^{2}$SLAM is displayed in Fig.~\ref{fig:archi}: it is based on a careful segmentation of critical and non-critical functions, either for ensuring RT processing or accuracy.  The next sections detail each thread and explain how computation is optimized to ensure getting high performance while respecting RT constraints.

\textbf{Assumptions.}  In this work, we consider calibrated optical systems, where both the intrinsic and the extrinsic parameters are known.  The pinhole camera model is used and both radial-tangential and fisheye distortion models are supported.  Unlike most of current VSLAM methods, we do not apply any rectification or undistortion to the images as those tends to crop the available field of view. This choice simply leads to storing additional variables as explained below but, in return, allows to fully exploit the acquired images, proving to be useful for wide field of view cameras.

\textbf{Notations.}  In the following, camera's pose at timestamp $i$ are represented as rigid body transformation $\mathbf{T_{wc_{i}}} \in \SE3$, which can transform the $k$-th 3D map point  $\boldsymbol{\lambda}^{w}_{k} \in \mathbb{R}^{3}$ expressed in the world frame $w$ to the current camera's frame: $\boldsymbol{\lambda}^{i}_{k} = \mathbf{T_{wc_{i}}}^{-1} \odot \boldsymbol{\lambda}^{w}_{k} = \mathbf{T_{c_{i}w}} \odot \boldsymbol{\lambda}^{w}_{k}$.

We denote the undistorded camera projection model as $\boldsymbol{\pi} : \mathbb{R}^{3} \mapsto \mathbb{R}^{2}$. Projection of a 3D map point to its undistorted pixel projection in the camera's image plane at instant $i$ writes: $\mathbf{x}_{ik} = \boldsymbol{\pi} \left(\mathbf{T_{iw}} \odot \boldsymbol{\lambda}^{w}_{k}  \right)$. In the stereo case, $\boldsymbol{\pi'}$ denotes the right camera projection and the undistorded right image is then given by:  $\mathbf{x}'_{ik} = \boldsymbol{\pi}' \left(\mathbf{T_{rl}} \mathbf{T_{iw}} \odot \boldsymbol{\lambda}^{w}_{k}  \right)$, where $\mathbf{T_{rl}} \in \SE3$ represents the rigid body transformation between the left and right camera.  We further consider the inverse projection, $\boldsymbol{\pi}^{-1}:\mathbb{R}^{2} \mapsto \mathbb{R}^{3}$, which projects undistorted keypoints to normalized coordinates.
When detecting or tracking a keypoint at some raw (distorded) pixel position, its undistorded position is immediately computed and stored together with raw coordinates. 

\section{The Visual Front-End}\label{sec:front}

The front-end thread is responsible for estimating the pose of the camera in real-time, i.e. at the camera's frame rate. More precisely, we define the front-end tasks as follows: image pre-processing, keypoint tracking, outlier filtering, pose estimation and keyframe creation triggering. The front-end pipeline is fully monocular, limiting all its operations to frames provided by the left camera with stereo setups.

\textbf{Image Pre-Processing.}  At the reception of a new image, we apply a contrast enhancement by means of CLAHE \cite{clahe94} which both increases the dynamic range and limits the intensity changes due to exposure adaptation.

\textbf{Keypoint Tracking.}  Keypoint tracking is performed by means of a guided coarse-to-fine optical flow method.  Keypoints are tracked individually using a pyramidal implementation of the inverse compositional Lucas-Kanade (LK) algorithm \cite{baker2004lucas} with a $9 \times 9$ pixel window and a pyramid scale factor of 2. The tracking process depends on the nature of the keypoint.
For 2D keypoints  -- \ie those with no prior information on their real 3D position -- , the initial position is set to their position in the previous frame and we use a four-levels pyramid.  For a 3D keypoint $k$ -- \ie already triangulated -- , its initial position is computed thanks to the projection model $\boldsymbol{\pi}$, its 3D position $\boldsymbol{\lambda}^{w}_k$ and the prediction of the current pose, $\mathbf{T_{wc_i}}$, using a constant velocity motion model.  We limit the minimization process to the first two levels of the image pyramid (\ie the ones with highest resolution).  Keypoints that we fail to track in this step are then searched for on a four-levels pyramid, along with 2D keypoints.
This two-stage process reduces the tracking run time while being robust to pose prediction errors or inaccurate 3D map points. We further avoid tracking errors by applying backward tracking only to the first level of the pyramid and removing keypoints that are more than 0.5 pixels away from their original position.  Tracked keypoints are finally updated  and their undistorted coordinates $\mathbf{x}_{ik}$ are computed from the known camera's intrinsic calibration.

\textbf{Outlier filtering.}  To eliminate outliers that can still occur in the tracking process, we apply RANSAC filtering based on the epipolar constraint. While many works use the Fundamental Matrix (FM) for this operation \cite{hartley2003multiple}, we prefer to use here the Essential Matrix (EM) as, in the case of VSLAM, camera's calibration is known.  This allows a faster RANSAC as only 5 correspondences are required in this case instead of the 7 or 8 usually required for the FM.  Furthermore, the estimation of an EM is more robust to planar scenes, which can lead to a degenerate FM \cite{nister2004efficient}.  To improve the RANSAC filtering efficiency, we estimate the EM from 3D keypoints only (which are more likely to be reliable than 2D ones) and then use it to filter non-consistent 2D keypoints. This filtering step ensures that there are virtually no outliers left before estimating the camera pose.

\textbf{Pose Estimation.}  Pose estimation is then performed by minimization of the 3D keypoints reprojection errors using a robust Huber cost function \cite{hartley2003multiple} $\left \| \cdot \right \|^{\varphi}$:
\begin{equation}
    \mathbf{T_{wc_i}^{*}} = \argmin_{\mathbf{T_{wc_i}}} \sum_{k} \left \| \mathbf{x}_{ik} - \boldsymbol{\pi} \left(\mathbf{T_{c_{i}w}} \odot \boldsymbol{\lambda}^{w}_{k}  \right) \right \|^{\varphi}_{\boldsymbol{\Sigma}_{ik}}
    \label{eq:pnp}
\end{equation}

where $\boldsymbol{\Sigma}_{ik}$ is the covariance associated to $\mathbf{x}_{ik}$.

This nonlinear optimization is performed with the Levenberg-Marquardt algorithm. The initial guess is of primary importance. We usually start from the pose predicted from the constant velocity motion model. 
However, it can happen that this prediction is wrong because the model of motion we are using is inadequate. Yet, the 3D keypoint tracking offers us a mean to detect such situations. More precisely, if the first step of the tracking, which starts from positions predicted thanks to the predicted pose results in less than half of the keypoints successfully tracked, we reject the predicted pose. In such case, we perform a new pose prediction using a P3P RANSAC \cite{kneip2011p3p} before applying eq.(\ref{eq:pnp}). At the end of the optimization, outliers are filtered using a $\chi^2$ test at 95\% on the reprojection errors \cite{mur2017orb}. 

\textbf{Keyframe Creation.}  Finally, the front-end thread is in charge of new keyframes creation.  Mainly, if the number of tracked 3D keypoints w.r.t. the last keyframe gets under a threshold (less than 85\% keypoints tracked) or if a significant parallax is detected (an average of 15 pixels of unrotated keypoints motion), a new keyframe is created. Detection of new keypoints is performed with a grid strategy, using cells of $35 \times 35$ pixels.  We process empty cells by selecting the point with the highest Shi-Tomasi score \cite{jianbo1994good} within the cell, then computing a subpixel refinement of its position. BRIEF descriptors \cite{calonder2011brief} are then computed for all (previously tracked or newly detected) keypoints. At this point, the front-end triggers the mapping thread to further process the created keyframe and continues its operations over the next frame.

\textbf{Initialization.}  Initialization is straightforward in stereo VSLAM as keypoints can be triangulated thanks to the known extrinsic calibration of the stereo rig. In the monocular case, we robustly compute an Essential Matrix $\mathbf{E}$ between the first two created keyframes. We then extract a relative pose from $\mathbf{E}$, choose an arbitrary scale and set the pose of the current keyframe. The extracted pose is then used by the mapping thread to initialize the 3D map.

\section{Mapping Thread}\label{sec:map}

The mapping thread is in charge of processing every new keyframe to create new 3D map points by triangulation and to track the current local map in order to minimize drift.  These two tasks do not have the same priority though. We give a higher priority to triangulation, which is critical for keeping accurate pose estimation in the front-end.  In practice, triangulation is applied first, then the local map tracking operation is executed and aborted if a new keyframe is available.

\textbf{Stereo Matching.}  
With stereo setups, a stereo matching step is applied.  Here, we follow the same coarse-to-fine optical flow strategy already used for keypoint tracking.  Both 2D and 3D keypoints are tracked in the right view (\ie non-triangulated and already triangulated keypoints) for both triangulation and the creation of additional stereo constraints in future BA steps.  This method is efficient but, due to the small basin of convergence of the LK equations, the initial guess is of primary importance.  For 3D keypoints, we simply project them in the right view using the estimated keyframe's pose.  For 2D keypoints, we initialise our guess using the depth of neighboring 3D keypoints if at least 3 can be found in the surrounding cells or use the left image pixel position otherwise.
Found stereo matches are then filtered by removing matches whose undistorted coordinates are away by more than 2 pixels from their corresponding epipolar line.

\textbf{Temporal Triangulation.}  Temporal triangulation is of course essential in the monocular case, as it is the only way to initialize a 3D map. However, we've found useful to also apply it in the stereo case, for keypoints correctly tracked until the current keyframe, but for which no stereo match could be found. The benefit is twofold. First, the 3D points created here will be useful for pose estimation as they are very likely to be tracked correctly in the following images.  Second, correct tracking will lead to an accurate 3D position, hence a good initial estimate for the next stereo matching step.  All the map points successfully triangulated are then immediately used by the front-end for localization purpose, while their 3D position will later be refined through BA.

\textbf{Local Map Tracking.} The local map denotes the set of 3D map points observed either by the current keyframe $K_i$ or one of its covisible keyframe $K_j$ (\ie a keyframe with at least one shared observation). The goal of the local map tracking is to find out if 3D map points belonging to this local map and not observed in $K_i$ can be matched to keypoints of $K_i$. Such "re-tracking" operations can be considered as elementary loop closures, limiting the accumulation of drift.

Any such 3D point whose projection onto $K_i$ is less than 2 pixels away from a keypoint defines a candidate match. 
As a 3D map point might be associated to several descriptors $\{\mathbf{d_j}\}$, we compute a distance between the keypoint candidate's descriptor $\mathbf{d_i}$ and every descriptor $\mathbf{d} \in \{\mathbf{d_j}\}$.  The candidate is finally accepted if the lowest computed distance is under a threshold.  This strategy increases the chance of correctly matching a 3D point to a keypoint as descriptors associated to a 3D map point are the ones extracted in the keyframes observing it, thus providing more robustness to evolution of appearance and viewpoint changes while still being extremely fast to compute thanks to the binary property of BRIEF.
Successful matches leads to the addition of new links in the covisibility graphs of the impacted keyframes, which will benefit to the upcoming local BA.

\section{State Optimization Thread}\label{sec:state}

The state optimization thread is in charge of running a local Bundle Adjustment to refine selected keyframes' poses and 3D map point positions. It additionally filters redundant keyframes to limit future local BA runtime.

\textbf{Local Bundle Adjustment.}  The local BA limits the drift coming from the visual measurements' noise by applying a multi-view optimization over the local map.  A classical choice is to refine only the poses of the most recent keyframes. 
In contrast, we follow ORB-SLAM approach by taking advantage of the covisibility graphs and include in the BA every keyframe that shares at least 25 common observations with the most recent keyframe $K_i$.  Every 3D map point observed by any of those keyframes is also included in the BA.  We denote the full set of parameters to optimize (poses and 3D map points) as $\boldsymbol{\zeta_i}$. Keyframes not in $\boldsymbol{\zeta_i}$ but observing a map point in $\boldsymbol{\zeta_i}$ are further added as fixed constraints for better conditioning.

3D map points are parameterized as anchored points with an inverse depth \cite{civera2008inverse, qin2018vins}, that is a 3D map point $\boldsymbol{\lambda}^{w}_{k}$ is defined by its undistorted pixel position $\mathbf{x}_{\alpha k}$ in its first observing keyframe $\mathbf{T_{w \alpha}}$ and its inverse depth $\gamma_k$ relative to this keyframe:  
\begin{gather}
    \boldsymbol{\lambda}^{\alpha}_{k} =  \frac{1}{\gamma_k} \boldsymbol{\pi}^{-1} \left( \mathbf{x}_{\alpha k}  \right) \\
    \boldsymbol{\lambda}^{w}_{k} = \mathbf{T_{w \alpha}} \odot \boldsymbol{\lambda}^{\alpha}_{k}
\end{gather}

Therefore, we limit the complexity of the BA by only having one d.o.f. to optimize per 3D map point.

The BA cost function is made of a collection of 2D reprojection errors, with, in the stereo case, additional terms corresponding to the right camera observation: 

\begin{equation}
\begin{split}
    \boldsymbol{\zeta_i}^{*} = \argmin_{\boldsymbol{\zeta_i}}  \sum_{j \in \mathfrak{K}_i} \sum_{k \in \mathfrak{L}_j} \left \| \mathbf{x}_{jk} - \boldsymbol{\pi} \left( \mathbf{T_{jw}} \cdot \mathbf{T_{w \alpha}}  \odot \boldsymbol{\lambda}^{\alpha}_{k} \right)  \right \|^{\varphi}_{\boldsymbol{\Sigma}_{jk}}  \\
    + \beta_j \cdot \left \| \mathbf{x'}_{jk} - \boldsymbol{\pi'} \left( \mathbf{T_{rl}} \cdot \mathbf{T_{jw}} \cdot \mathbf{T_{w \alpha}}  \odot \boldsymbol{\lambda}^{\alpha}_{k} \right)  \right \|^{\varphi}_{\boldsymbol{\Sigma'}_{jk}} 
\end{split}
\end{equation}

\noindent where $\mathfrak{K}_i$ is the full set of keyframes contributing to the BA, $\mathfrak{L}_j$ is the set of map points in $\boldsymbol{\zeta_i}$ observed by $K_j$ and $\beta_j$ is set to 1 if a map point is also observed in the right camera and to 0 otherwise.

This nonlinear least-squares optimization is solved on-manifold \cite{ManifoldOptAbsil2007} with the Levenberg-Marquardt algorithm.  Outliers marked by the robust Huber function are then removed as done after solving eq.(\ref{eq:pnp}) and optimized states are updated.

\textbf{Keyframes Filtering.}  Next, we apply a filtering step to remove any redundant keyframe.  Exploring the current keyframe's covisibility graph, we remove any keyframe for which at least 95\% of their observed 3D map points are already observed by at least 4 other keyframes. This limits the growth in the number of states included in the next BA without giving up on accuracy as removal is only applied to merely informative keyframes.  For 3D points anchored in a keyframe getting filtered, we simply switch their anchor for their second observing keyframe.

\section{Online Bag-of-Words based Loop Closer}\label{sec:bowlc}

While already being beneficial in small and medium environments, Loop Closing (LC) becomes a key feature for accurate long-term localization in large-scale maps.  The LC thread is hence responsible for detecting such loops and conducting relocalization, that is correct both the current pose estimates and the estimated trajectory between the current frame and the passed keyframe where LC has been detected. The challenge is to do these operations as often (therefore as quickly) as possible while rejecting false detections that would significantly corrupt the trajectory.

\textbf{Online Bag-of-Words.} Bag-of-Words (BoW) have proved to be very efficient for fast LC detection.  The vocabulary tree defined within the BoW along with the "term frequency - inverse document frequency" (tf-idf) and the inverted index allows fast computation of similarity scores between different keyframes.  Yet, most of the SLAM algorithms make use of offline trained BoW, hence dependent and biased by the training database.   
Another strategy is to build the vocabulary tree online \cite{angeli2008fast,nicosevici2012automatic}, using the images acquired so far. The idea here is to create a vocabulary tree that fits the current environment, hence avoiding the potentially underfitting issues of offline trained ones. This is the approach chosen in \ov2slam using a modified version of iBoW-LCD \cite{Garcia-Fidalgo2018ibow-lcd} to perform LC detection.  
We stick to the original implementation to find LC candidates but perform our own processing to assess the correctness of this candidate. 

\textbf{Keyframe's Pre-Processing.}  Upon reception of a new keyframe, we perform a feature extraction step.  As we do not track many keypoints for the purpose of localization (usually around 200), we extract additional features before updating the vocabulary tree and searching for a LC candidate.  Features are extracted with the FAST detector \cite{rosten2006machine} and BRIEF descriptor, keeping the best 300 features.  The vocabulary tree is then updated using both the SLAM features and these new features, and the resulting BoW signature is used to compute a similarity score with previous keyframes.

\textbf{Loop Candidate Processing.} If a good LC candidate is found, we first ensure that it is not a false positive.  Given the current keyframe $K_i$ and the candidate keyframe $K_{lc}$, we first apply a k-Nearest-Neighbor Brute-Force matching between all the descriptors in both keyframes.   Ambiguous matches are first filtered by a classical ratio test \cite{lowe2004distinctive}.  We then compute an Essential Matrix within a RANSAC scheme in order to keep only the matches satisfying the epipolar geometry.  Using the remaining inliers, we compute a pose hypothesis for $K_i$ through a P3P-RANSAC method \cite{kneip2011p3p} with the remaining 3D map points observed by $K_{lc}$.  If a reliable pose is found, determined by the resulting number of inliers, we get the local map of $K_{lc}$ and search for additional matches in $K_{i}$, projecting the 3D map points using the P3P computed pose.  The hypothesis pose is then refined based on all the matching found so far using eq.(\ref{eq:pnp}) and a final filtering step is performed based on the outliers detected by the Huber robust cost function. 

\textbf{Pose Graph Optimization.} Finally, if at least 30 inliers remain, we validate the LC detection and perform a Pose Graph Optimization (PGO) to correct the full trajectory.  PGO is performed over the trajectory's part starting from $K_{lc}$ up to $K_i$ and seeks to minimize the relative pose errors between consecutive keyframes given their initial relative pose (i.e. the ones before LC detection) and the newly estimated pose for $K_i$.  A Levenberg-Marquardt optimization is performed over the $\mathbb{SE}(3)$ keyframes' poses and we use a second-order approximation of the Campbell-Baker-Haussdorf formula \cite{barfoot2017state} to compute their respective Jacobian on $\mathfrak{se}(3)$.
 
As new keyframes might have been created since the start of the current LC detection, we propagate the pose graph corrections to the most recent part of the trajectory, updating keyframes using their previous relative pose and the corrected pose of $K_i$.  3D map points matched between $K_i$ and $K_{lc}$ are then merged and all the 3D map points first observed by any corrected keyframe are updated through a forward-backward projection to align the 3D map w.r.t. to the LC corrections:

\begin{equation}
    {\boldsymbol{\lambda}^{w}_{k}}^{*} = \mathbf{T^{new}_{w\alpha}} \cdot  \mathbf{T^{old}_{\alpha w}} \odot \boldsymbol{\lambda}^{w}_{k}
    \label{eq:fb-proj}
\end{equation}

\textbf{Loose Bundle Adjustment.} Once PGO is done, we apply a "loose-BA", which means that we apply a BA optimization only to the part of the map impacted by the LC corrections (\ie all the 3D map points observed by corrected keyframes and all the keyframes observing these 3D points).  Thus, we limit the overload of applying a full-BA (as done in ORB-SLAM) by optimizing over a subset of keyframes and 3D map points. This strategy leads to a largely decreased run-time for LC involving recent keyframes.
As loose-BA optimization might still take a few seconds to be performed, new keyframes and map points added meanwhile have to be updated as well to keep the trajectory consistent. It is done in the same way as for PGO, propagating the correction on the keyframes poses using their previous relative pose and updating the 3D map points they observe through eq.(\ref{eq:fb-proj}).

\section{Experiments}\label{sec:expe}

\textbf{Implementation.}  The proposed method has been developed in C++, runs on CPU only and relies on the ROS middleware for data reading and visualization.  Feature-related operations are performed using the OpenCV library, multi-view geometry computations are run with OpenGV \cite{kneip2014opengv} and nonlinear optimization is applied using the Ceres library.  In all experiments, the covariance associated to visual measurements $\mathbf{x_{ik}}$ is set as $\boldsymbol{\Sigma_{ik}} = \mathbf{I_{2\times2}}$.
In addition to the settings detailed throughout the paper, we propose a lightweight configuration called \fov2slam, able to process sequences at hundreds of Hertz.  To reach such performances, we modify the settings of \ov2slam as follows: we disable LC, we use FAST \cite{rosten2006machine} for keypoint detection and increase the grid's cell size to 50 $\times$ 50 pixels.  

We run all our experiments on a high-end laptop equipped with an Intel Xeon (8 threads @3.00 GHz / 32 Gb RAM) except for EuRoC dataset in monocular setup in which an i5 (4 threads @2.20 GHz / 8 Gb RAM) laptop is used.  A run time analysis is performed on sequence MH03 of EuRoC and Table \ref{tab:timings} shows the average timings for the most important functions of the algorithm obtained on both architectures with both \ov2slam and \fov2slam. The full front-end thread computation time is upper bounded by the sum of the ``Front-End Tracking" and ``Keyframe Creation", showing that the usual acquisition rates of 20-30 Hz are easily supported by either architecture and can even go up to several hundreds of Hertz when running \fov2slam.  
The configuration files used for running all the following experiments are included in the open-source repository.

\begin{table}[!h]
\begin{adjustbox}{width=\linewidth}
\begin{tabular}{@{}ccccccc@{}}
\toprule
      &                                   & \multicolumn{5}{c}{Timings (ms)}      \\
\cmidrule{3-7}
      &                                   & Front-End & Keyframe & Mapping & Local & LC  \\
      &                                   & Tracking & Creation & Thread & BA & Detection \\
\midrule
\midrule
      
Intel Xeon 8T@3.0 GHz    & \ov2slam w. LC & 8.01        & 6.82       & 24.90    & 73.36        & 42.76        \\
 32 Gb RAM                                   & \fov2slam  & 6.05        & 5.22       & 11.37    & 18.68        & -            \\
 
\midrule

Intel i5 4T@2.20 GHz             & \ov2slam w. LC & 16.49       & 14.02      & 43.87    & 113.35       & 80.16        \\
8 Gb RAM                                        & \fov2slam  & 7.61        & 7.07       & 12.63    & 27.10        & -            \\
\bottomrule
\end{tabular}
\end{adjustbox}
\caption{\label{tab:timings} Timing of the main functions of \ov2slam in stereo mode.}
\end{table}

\textbf{Datasets.}  We evaluate \ov2slam on the widely used benchmarking datasets EuRoC \cite{burri2016euroc} and KITTI \cite{geiger2012we} as well as on the very recent dataset TartanAir \cite{wang2020tartanair}.  The EuRoC dataset is dedicated to localization of Micro-Aerial-Vehicles (MAV) in small to medium scaled environments, 6 sequences being acquired in two small rooms (3 seq. per room) and 5 sequences acquired in a factory hall.  For each one of these environments, the sequences were acquired with an increasing difficulty by performing more and more aggressive motions with the MAV.  The KITTI dataset is dedicated to autonomous car driving and is the main dataset used to evaluate the capacitiy of SLAM methods in handling large-scale environments, with some trajectories being several kilometers long.  In opposition to these real world datasets, the TartanAir dataset is a photo-realistic one created with Unreal engine.  It is meant for pushing the limit of VSLAM by proposing sequences with extremely diverse environments, motion patterns and lightning conditions (day / night, sunny / rainy, ...).  Either the ATE or the RPE metrics \cite{sturm2012benchmark} are used for evaluation on these datasets, depending on what has been classically used in the literature.   Figures of the resulting trajectories are available in the \hyperref[sec:supp-material]{\textbf{supplementary material}}.

\subsection{EuRoC MAV Dataset} \label{sec:expe_euroc}

\begin{table*}[!t]
\centering
\begin{adjustbox}{width=0.9\linewidth}
\begin{tabular}{@{}cccccccccc@{}}
\toprule

&            &   \multicolumn{3}{c}{Visual SLAM w. LC}       &     \multicolumn{2}{c}{Visual SLAM no LC}      &     \multicolumn{3}{c}{Visual-Inertial SLAM}  \\

\cmidrule(l){3-5}
\cmidrule(l){6-7}
\cmidrule(l){8-10}

Seq.   & Length (m) & ORB-SLAM (not RT) & ORB-SLAM & \ov2slam           & SVO            & \ov2slam              & OKVIS              & Vins-Fusion & BASALT         \\
\midrule
\midrule
MH 01  &   79.84    &   \itbf{0.04}   &   0.14      &   \bblue{0.04}    &   \bblue{0.04} &   0.05                &   0.23            &   0.24      &   0.08          \\
MH 02  &   72.75    &   \itbf{0.02}   &   0.12      &   \bblue{0.04}    &   0.05         &   \bblue{0.04}        &   0.15            &   0.18      &   0.06          \\
MH 03  &   130.58~   &   \itbf{0.03}   &   ~0.31*     &   \bblue{0.04}  &   0.06         &   \textbf{0.05}       &   0.23            &   0.23      &   0.05          \\
MH 04  &   91.55    &   0.12           &   0.25      &   \bblue{0.06}   &   $\times$     &   \textbf{0.12}       &   0.32            &   0.39      &   0.10          \\
MH 05  &   97.32    &   \itbf{0.06}   &   0.28      &   \bblue{0.07}    &   0.12         &   \textbf{0.10}       &   0.36            &   0.19      &   0.08          \\
V1 01  &   58.51    &   \itbf{0.04}   &   ~0.18*     &   \textbf{0.09}  &   \bblue{0.05} &   0.09                &   \blue{0.04}     &   0.10      &   \blue{0.04}   \\
V1 02  &   75.72    &   \itbf{0.02}   &   $\times$   &   \textbf{0.07}  &   \bblue{0.05} &   0.08                &   0.08            &   0.10      &   \blue{0.02}   \\
V1 03  &   78.77    &   \itbf{0.05}   &   $\times$   &   \bblue{0.09}   &   $\times$     &   \textbf{0.17}       &   0.13            &   0.11      &   \blue{0.03}   \\
V2 01  &   36.34    &   \itbf{0.04}   &   ~0.29*     &   \textbf{0.07}  &   \bblue{0.05} &   0.10                &   0.10            &   0.12      &   \blue{0.03}   \\
V2 02  &   83.01    &   \itbf{0.04}   &   $\times$   &   \bblue{0.06}   &   $\times$     &   \textbf{0.10}       &   0.17            &   0.10      &   \blue{0.02}   \\
\bottomrule
\end{tabular}
\end{adjustbox}
\caption{\label{tab:stereo_euroc} Comparison of ATE rmse (m) for stereo methods on EuRoC (20 Hz).  For RT VSLAM methods, best results are shown in bold blue and second best results in bold. If best, ORB-SLAM not RT results are shown in bold italic and VI-SLAM ones in blue.  * indicates frequent failures.}
\end{table*}

We first compare the stereo version of \ov2slam with the stereo VSLAM methods ORB-SLAM and SVO as well as the stereo VI-SLAM algorithms OKVIS \cite{leutenegger2015okvis}, Vins-Fusion \cite{qin2019a} and Basalt \cite{usenko2019visual} on EuRoC \cite{burri2016euroc} in Table \ref{tab:stereo_euroc}.  The results for the VI-SLAM methods are reported from \cite{usenko2019visual} and were obtained enforcing RT processing. Results for SVO are reported from the authors' evaluation (Table I in \cite{forster2016svo}) which were obtained in RT.  For ORB-SLAM, we both report the non RT results and the ones that we obtained when enforcing RT processing, keeping the same settings as those proposed by the authors.   
All the reported results are the median accuracy obtained over 5 runs and we do not use the sequence V2 03 because hundreds of images from the right camera are missing.  As the reader can see, running the stereo version of ORB-SLAM when enforcing RT leads to big drops in terms of accuracy -- being even more significant given the high-end laptop used for running the experiments.  Taking the sequences acquired in the factory hall (MH XX), we can see that \ov2slam outperforms all the RT methods, including the VI-SLAM ones.  Furthermore, even \ov2slam without LC manages to outperform all the methods on all but one sequence.  One can also notice that the performances in terms of accuracy are very close to ORB-SLAM not running in RT.  On the small room sequences (VX XX), both version of \ov2slam have comparable results to OKVIS and Vins-Fusion but SVO and Basalt mostly get better results.  This is mainly due to the fact that these rooms are extremely low-textured, making direct VSLAM methods more efficient and the use of an IMU can be beneficial on these sequences.

\begin{table}[!h]

\begin{adjustbox}{width=\linewidth}
\begin{tabular}{@{}ccccccc@{}}
\toprule

       &           \multicolumn{5}{c}{\fov2slam}        & \ov2slam no LC. \\ 

\cmidrule(l){2-6}
\cmidrule(l){7-7}

Seq.    &    100 Hz   & 150 Hz   &   200 Hz   &  300 Hz   &   400 Hz  &  20 Hz   \\
\midrule
\midrule
MH 01   &   0.061  &   0.059   &   0.070     &  0.060    &  0.066    &  0.054  \\
MH 02   &   0.056  &   0.068   &   0.061     &  0.051    &  0.056    &  0.041  \\
MH 03   &   0.113  &   0.121   &   0.337     &  x        &  x        &  0.058  \\
MH 04   &   0.256  &   0.214   &   0.263     & ~0.233*   &  ~1.840*  &  0.116  \\
MH 05   &   0.165  &   0.165   &   0.192     & ~0.223*   &  ~1.420*    &  0.140   \\
V1 01   &   0.186  &   0.257   &   0.179     &  0.498    &  0.353    &  0.093  \\
V1 02   &   0.590  &   0.578   &   0.624     &  x        &  x        &  0.088  \\
V1 03   &   x      &   x       &   x         &  x        &  x        &  0.290   \\
V2 01   &   0.117  &   0.126   &   0.103     &  0.121    &  0.780    &  0.092  \\
V2 02   &   0.271  &   0.336   &   x         &  x        &  x        &  0.098  \\

\bottomrule
\end{tabular}
\end{adjustbox}
\caption{\label{tab:fast_euroc} Comparison of ATE rmse (m) for stereo \fov2slam when run on EuRoC at high rates.  * indicates frequent failures.}
\end{table}

In order to highlight the RT performances of \fov2slam, we run it on EuRoC, playing the sequences at higher rates than the real one.  More specifically, we run the sequences from 5 to 20 times the original rate.  In this experiment, the factory sequences, we start running the sequences right after the aggressive motions performed for the means of VI-SLAM initialization 
.  The obtained results are reported in Table \ref{tab:fast_euroc}.  As one can see, we manage to get impressive results, reaching almost the same results than when processing the sequences in real-time (\ie at 20 Hz) on the easiest sequences at rates up to 400 Hz.  Moreover, comparing Table \ref{tab:fast_euroc} to Table \ref{tab:stereo_euroc}, we can observe that the results obtained up to 200 Hz are very close to the ones obtained with ORB-SLAM at 20 Hz.

\begin{table}[!b]
\begin{adjustbox}{width=\linewidth}
\begin{tabular}{@{}cccccc@{}}
\toprule

        & Not Real-Time             &  \multicolumn{4}{c}{Real-Time}           \\ 

\cmidrule(l){2-2}
\cmidrule(l){3-6} 

Seq.       & ORB-SLAM no LC  & ORB-SLAM no LC   &   SVO           & DSO              & \ov2slam no LC   \\
\midrule
\midrule
MH 01      &  \itbf{0.02}    &    0.61          &   0.06          & \bblue{0.05}     & \bblue{0.05}    \\
MH 02      &  0.03           &    0.72          &   0.07          & \textbf{0.05}    & \bblue{0.03}    \\
MH 03      &  \itbf{0.03}    &    1.70          &   $\times$      & \textbf{0.26}    & \bblue{0.06}    \\
MH 04      &  0.22           &    6.32          &   0.40          & \textbf{0.24}    & \bblue{0.17}    \\
MH 05      &  0.71           &    5.66          &   $\times$      & \bblue{0.15}     & \textbf{0.16}   \\
V1 01      &  0.16           &    1.35          & \bblue{0.05}    & 0.47             & \textbf{0.09}   \\
V1 02      &  0.18           &    0.58          &  $\times$       & \bblue{0.10}     & \textbf{0.22}   \\
V1 03      &  0.78           &   \textbf{0.63}  &  $\times$       & 0.66             & \bblue{0.58}    \\
V2 01      &  \itbf{0.02}    &    0.53          &  $\times$       & \bblue{0.05}     & \textbf{0.06}   \\
V2 02      &  0.21           &    0.68          &  $\times$       & \bblue{0.19}     & \textbf{0.33}   \\
V2 03      &  1.25           &    \bblue{1.06}  &  $\times$       & \textbf{1.19}    &   $\times$      \\ 

\bottomrule
\end{tabular}
\end{adjustbox}
\caption{\label{tab:mono_euroc} Comparison of ATE rmse (m) for monocular methods on EuRoC (20 Hz).  For RT methods, best results are shown in bold blue and second best results in bold.  ORB-SLAM not RT results are shown in bold italic if best.  * indicates frequent failures..}
\end{table}

Finally, we compare the monocular version of \ov2slam to ORB-SLAM, SVO and DSO.  Neither DSO or SVO implements LC so we compare to the results reported in \cite{forster2016svo}, obtained without LC for ORB-SLAM and comparing ORB-SLAM with and without enforcing RT processing.  The reported results from \cite{forster2016svo} were obtained with an i7 @2.80 GHz architecture.  To be fair, we use the consumer-grade laptop (i5 @2.20 GHz) here to run \ov2slam.  The results are reported in Table \ref{tab:mono_euroc}.  Once again, \ov2slam mostly outperforms others methods on the factory sequences.  It also shows high robustness, handling almost all the sequences in the low-textured rooms with competitive accuracy.  DSO is here the closest method to \ov2slam in terms of performances.

\subsection{KITTI Dataset}

In this section, we evaluate \ov2slam on KITTI \cite{geiger2012we} and compare to ORB-SLAM on KITTI's train set.  Results are reported in Table \ref{tab:stereo_kitti} and highlights the fact that ORB-SLAM is highly impacted by the run-time requirement of a sequence, significantly decreasing its performances in terms of accuracy.  In opposition, we show that \ov2slam is barely impacted by the run-time requirements here and clearly outperforms ORB-SLAM when real-time is enforced.
We further report the results for open-sourced VSLAM methods on KITTI's online benchmark in Table \ref{tab:stereo_kitti_online} and show that, at the time of writing, \ov2slam is the best open-sourced method on this dataset.

\begin{table}[!t]
\centering
\begin{adjustbox}{width=\linewidth}
\begin{tabular}{@{}ccccccc@{}}
\toprule

     &        &  Not Real-Time &  \multicolumn{2}{c}{Real-Time}  & \multicolumn{2}{c}{0.5$\times$ Real-Time}  \\

\cmidrule(l){3-3} 
\cmidrule(l){4-5} 
\cmidrule(l){6-7}   

 Seq.  & Length (m)        & ORB-SLAM    & ORB-SLAM    & \ov2slam             & ORB-SLAM      &  \ov2slam           \\ 
\midrule \midrule
00 & 3724.2                & 1.3         & ~10.74*     & \textbf{1.17}        & 3.79          &  \textbf{1.0}       \\
01 & 2453.2                & \itbf{10.4} & 341.81      & \textbf{31.95}       & 71.62         &  \textbf{26.34}     \\
02 & 5067.2                & \itbf{5.7}  & $\times$    & \textbf{6.24}        & 34.26         &  \textbf{6.45}      \\
03 & 560.9                 & \itbf{0.6}  & 2.55        & \textbf{1.25}        & \textbf{0.58} &  1.28               \\
04 & 393.6                 & \itbf{0.2}  & ~5.76*      & \textbf{1.18}        & 3.86          &  \textbf{1.27}      \\
05 & 2205.6                & \itbf{0.8}  & 9.15        & \textbf{1.44}        & 3.70          &  \textbf{1.39}      \\
06 & 1232.9                & \itbf{0.8}  & $\times$    & \textbf{1.27}        & 4.92          &  \textbf{1.27}      \\
07 & 694.7                 & 0.5         & 1.07        & \textbf{0.37}        & 0.53          &  \textbf{0.36}      \\
08 & 3222.8                & 3.6         & ~5.41*      & \textbf{3.68}        & 3.81          &  \textbf{3.57}      \\
09 & 1705.1                & 3.2         & ~5.11*      & \textbf{1.59}        & 2.92          &  \textbf{1.58}      \\
10 & 919.5                 & 1.0         & ~2.91*      & \textbf{0.66}        & 1.11          &  \textbf{0.64}      \\ 
\bottomrule
\end{tabular}
\end{adjustbox}
\caption{\label{tab:stereo_kitti} Comparison ATE rmse (m) for the stereo versions of ORB-SLAM and \ov2slam on KITTI training sequences played RT (10 Hz) and half RT (5 Hz).  Best results are shown in bold and ORB-SLAM not RT results are shown in bold italic if best.  * indicates frequent failures.}
\end{table}

\begin{table}[!h]
\centering
\begin{tabular}{@{}ccc@{}}
\toprule

 Method & t$_{rel}$ (\%)   &    r$_{rel}$ (deg / m)    \\
  \midrule  \midrule
  \ov2slam w. LC  &    0.94     &    0.0023  \\
  \ov2slam no LC  &    0.98     &    0.0023   \\
  Vins-Fusion \cite{qin2019a}  &    1.09     &    0.0033   \\
  ORB-SLAM \cite{mur2017orb} &    1.15     &    0.0027   \\
  S-PTAM \cite{pire2017sptam}  &    1.19     &    0.0025   \\
  RTAB-Map \cite{labbe2019rtab} &    1.26     &    0.0026   \\
\bottomrule
\end{tabular}
\caption{\label{tab:stereo_kitti_online} Comparison of translational rmse (\%) and rotational rmse (deg/m) on KITTI's online benchmark for open-sourced stereo methods.}
\end{table}

\subsection{TartanAir Dataset}

This section details the results obtained on the TartanAir dataset \cite{wang2020tartanair}.  More specifically, we report the results obtained on the sequences used for the SLAM Challenge organized within the Visual SLAM workshop at CVPR 2020.  This challenge was divided into two tracks, one for monocular VSLAM and one for stereo VSLAM.   The results were averaged over 16 sequences, half considered as \textit{easy} and the other half as \textit{hard}.  The results are showed in Table \ref{tab:tartanair} and come both from the official benchmark website\footnote{\url{https://tinyurl.com/y3ggvvta}, \url{https://tinyurl.com/y3nh6yuo}} and from the post challenge technical discussion\footnote{\url{https://www.youtube.com/watch?v=jNhPD4oO6xA}}. As it can be seen, the best performing method used Colmap \cite{schonberger2016structure}, a Structure-from-Motion (SfM) library, along with deep features \cite{detone2018superpoint,sarlin2020superglue} for matching.  The second best performing method was based on Voldor \cite{min2020voldor}, a dense VSLAM methods that computes residual flow for pose estimation, along with Colmap to scale the estimated translation.  \ov2slam ranks 3rd on both the monocular and stereo tracks of this challenge and is actually the 1st if we only consider online methods, \ie~methods that only use the past and present to perform estimations.  The challenge's organizers have run ORB-SLAM on both tracks to provide a baseline (without enforcing RT) and, as one can see, the performances obtained with \ov2slam are dramatically better, highlighting its better robustness to challenging and very diverse environments, even without enabling the LC feature.  Furthermore, we have run \ov2slam at 20 Hz before submitting the estimated trajectories on these sequences and we can see that it compares very competitively to offline SfM methods while not having their run-time burden (the team which ranked 1st reports half an hour to process a sequence with 1000 images).

\begin{table}[!h]
\begin{adjustbox}{width=\linewidth}
\begin{tabular}{@{}cccccc@{}}
\toprule

                      &  \multicolumn{2}{c}{Monocular Track}  & \multicolumn{2}{c}{Stereo Track}  &   Online  \\ 
\cmidrule(l){2-3} \cmidrule(l){4-5} \cmidrule(l){6-6}
    Method  & ATE (m) & RPE (m) & ATE (m) & RPE (m) &  \\ 
\midrule \midrule
Colmap + SuperPoint + SuperGlue &  -              &  -          & 0.119           &  0.484  & X             \\
Colmap + SuperPoint + SIFT      &  0.34           &  0.449      & -               &  -      & X             \\
Colmap + Voldor                 &  0.44           &  1.273      & 0.177           &  0.570   & X             \\
\ov2slam w. LC                  &  -              &  -          & 0.182           &  1.025  & \checkmark    \\
\ov2slam no LC                  &  0.51           &  0.889      & 0.199           &  0.804  & \checkmark    \\
ORB-SLAM                        &  3.57           &  17.700~       & 1.640            &  2.900    & \checkmark    \\

\bottomrule
\end{tabular}
\end{adjustbox}
\caption{\label{tab:tartanair} Comparison of ATE and RPE (m) on TartanAir's online challenge for submitted methods in both the monocular and stereo tracks.  Methods are tagged as online if they process sequences sequentially.}
\end{table}

\section{CONCLUSIONS}

In this work, we have presented \ov2slam a complete VSLAM algorithm that aims at closing the gap between accuracy, robustness and RT capability. \ov2slam is also design to be versatile and we have successfully used it with terrestrial, aerial and pedestrian setups in very different environments (indoor, outdoor).  We have detailed the careful design of its architecture, allowing to respect the RT constraints required by real-world applications without sacrificing accuracy.  It further integrates an online BoW methods for very efficient loop-closure detection.  \ov2slam has been evaluated on several datasets, both in monocular and stereo setups, and show state-of-the-art performances.  By releasing its source code, we hope that it could make an interesting  ready-to-use VSLAM research platform.




\bibliographystyle{ieeetr} 
\bibliography{biblio.bib}


\pagebreak

\section{Supplementary Material} \label{sec:supp-material}

This report contains figures of trajectories estimated in the experiment section of \ov2slam paper.  We both provide results obtained on the training sequences of the KITTI dataset \cite{geiger2012we} and on the EuRoC dataset \cite{burri2016euroc}.

\section{EuRoC Experiments}

\begin{figure}[!ht]
    \begin{subfigure}[b]{0.475\linewidth}
        \centering \includegraphics[width=0.975\linewidth]{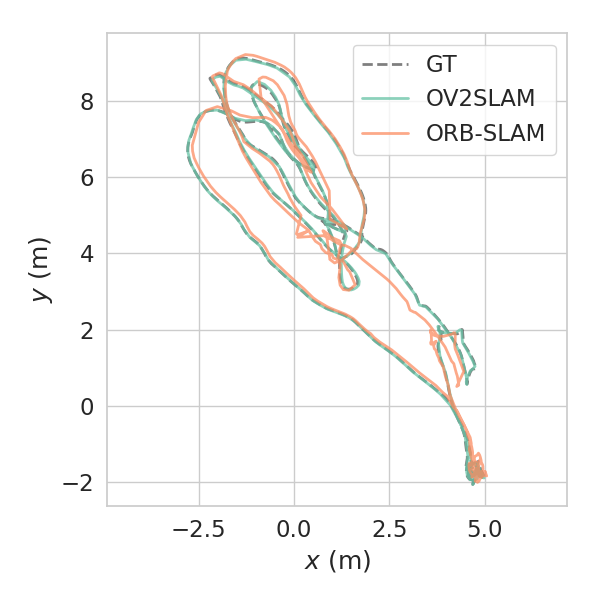}
        \captionsetup{justification=centering}
        \caption{EuRoC MH01}\label{fig:colmap_1}
    \end{subfigure}
    \begin{subfigure}[b]{0.475\linewidth}
        \centering \includegraphics[width=0.975\linewidth]{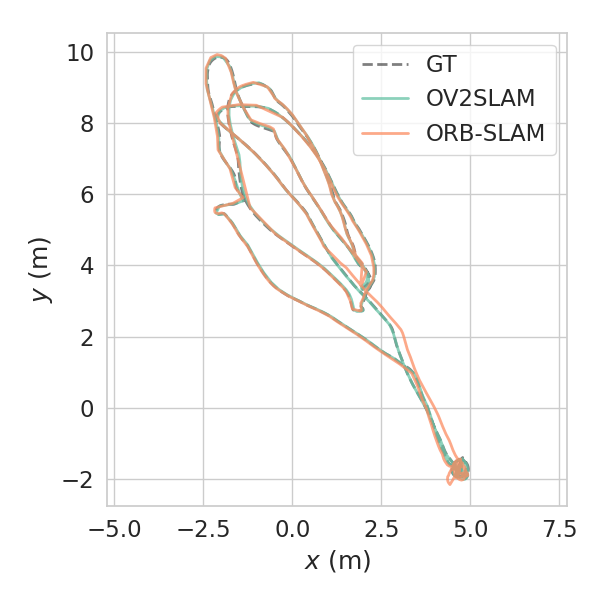}
        \captionsetup{justification=centering}
        \caption{EuRoC MH02}\label{fig:colmap_1}
    \end{subfigure}
    
    \begin{subfigure}[b]{0.475\linewidth}
        \centering \includegraphics[width=0.975\linewidth]{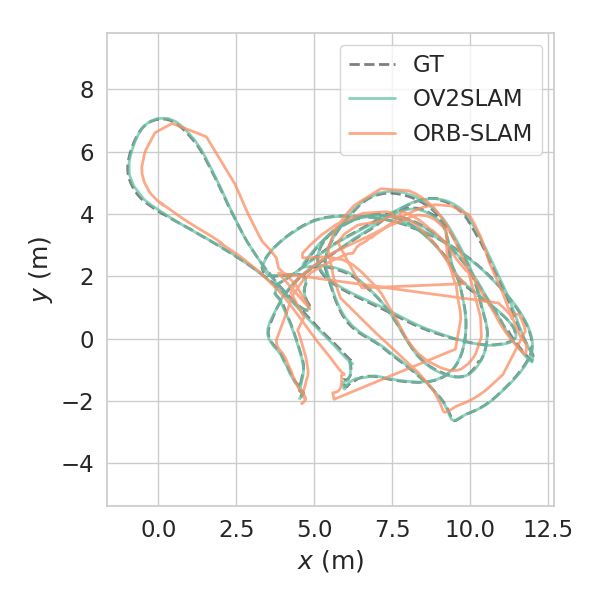}
        \captionsetup{justification=centering}
        \caption{EuRoC MH03}\label{fig:colmap_1}
    \end{subfigure}
    \begin{subfigure}[b]{0.475\linewidth}
        \centering \includegraphics[width=0.975\linewidth]{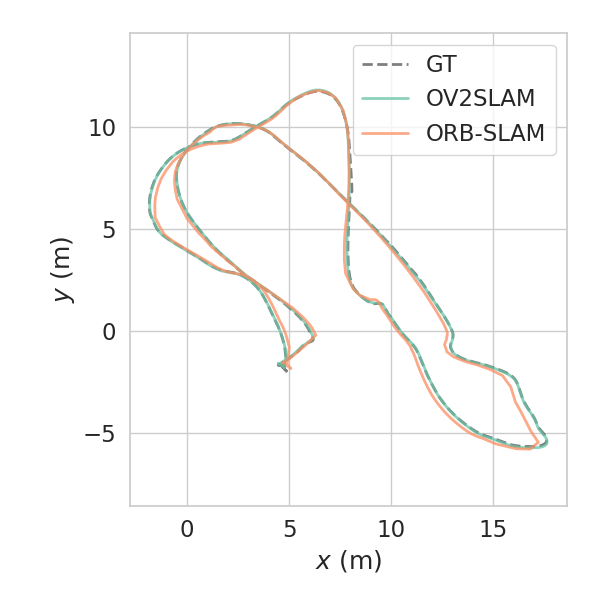}
        \captionsetup{justification=centering}
        \caption{EuRoC MH04}\label{fig:colmap_1}
    \end{subfigure}
    
    \begin{subfigure}[b]{0.475\linewidth}
        \centering \includegraphics[width=0.975\linewidth]{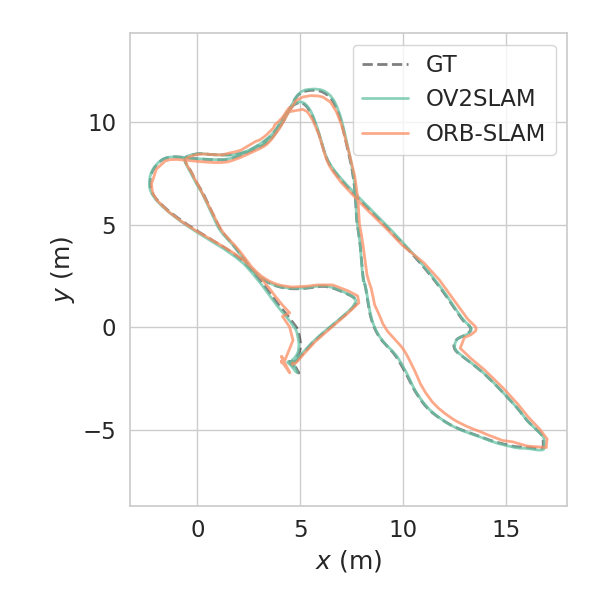}
        \captionsetup{justification=centering}
        \caption{EuRoC MH05}\label{fig:colmap_1}
    \end{subfigure}
    
    \captionsetup{justification=centering}
    \caption{Trajectories estimated in Real-Time with stereo \ov2slam and ORB-SLAM on EuRoC Machine Hall (MHXX) sequences.}
    
    \label{fig:euroc-stereo-mh}
\end{figure}

\begin{figure}[!ht]
    \begin{subfigure}[b]{0.475\linewidth}
        \centering \includegraphics[width=0.975\linewidth]{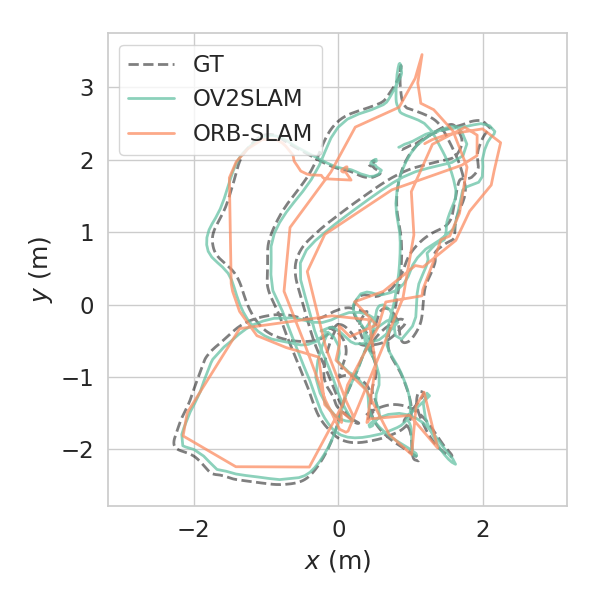}
        \captionsetup{justification=centering}
        \caption{EuRoC V1 01}\label{fig:colmap_1}
    \end{subfigure}
    \begin{subfigure}[b]{0.475\linewidth}
        \centering \includegraphics[width=0.975\linewidth]{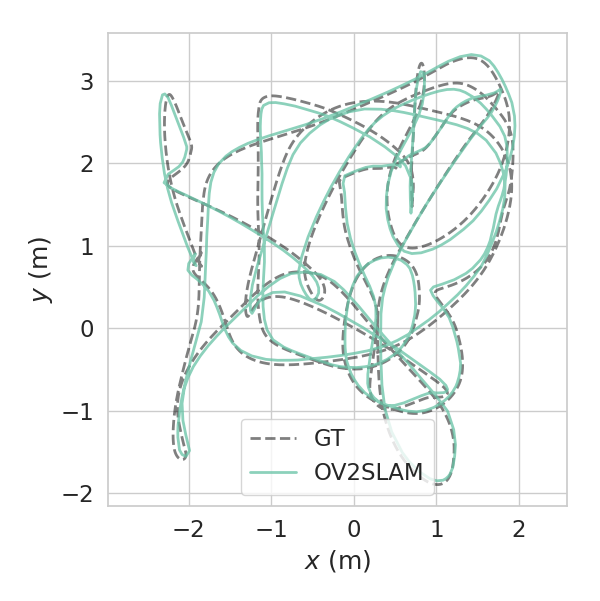}
        \captionsetup{justification=centering}
        \caption{EuRoC V1 02}\label{fig:colmap_1}
    \end{subfigure}
    
    \begin{subfigure}[b]{0.475\linewidth}
        \centering \includegraphics[width=0.975\linewidth]{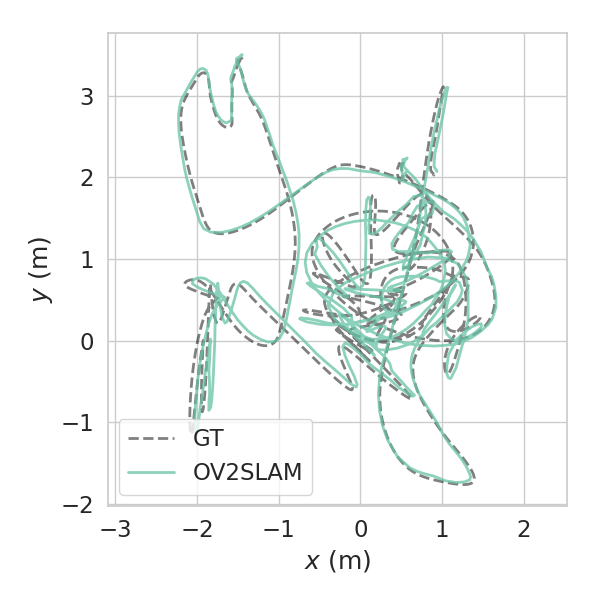}
        \captionsetup{justification=centering}
        \caption{EuRoC V1 03}\label{fig:colmap_1}
    \end{subfigure}
    \begin{subfigure}[b]{0.475\linewidth}
        \centering \includegraphics[width=0.975\linewidth]{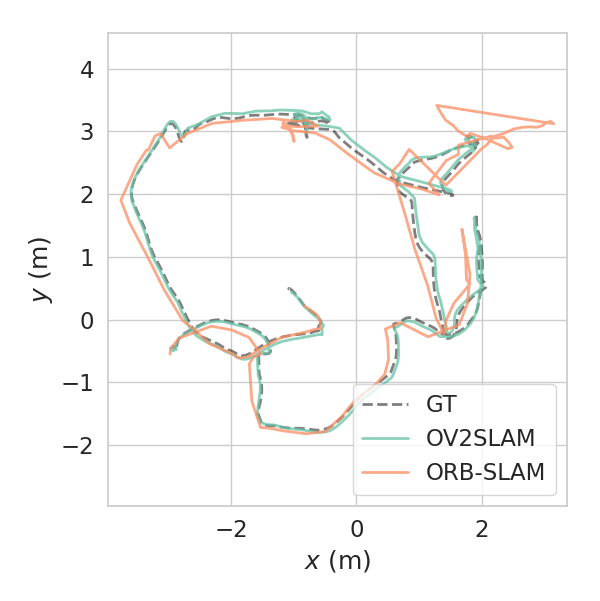}
        \captionsetup{justification=centering}
        \caption{EuRoC V2 01}\label{fig:colmap_1}
    \end{subfigure}
    
    \begin{subfigure}[b]{0.475\linewidth}
        \centering \includegraphics[width=0.975\linewidth]{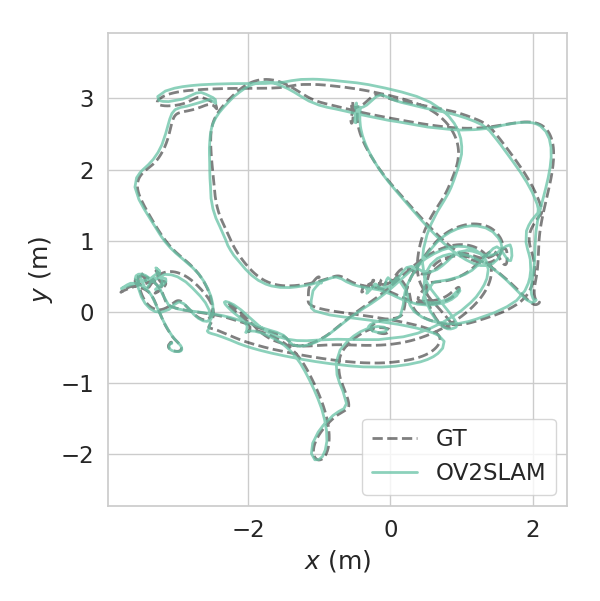}
        \captionsetup{justification=centering}
        \caption{EuRoC V2 02}\label{fig:colmap_1}
    \end{subfigure}
    
    \captionsetup{justification=centering}
    \caption{Trajectories estimated in Real-Time with stereo \ov2slam and ORB-SLAM on EuRoC Vicon Room (VX-XX) sequences.}
    
    \label{fig:euroc-stereo-vx}
\end{figure}

We compare the stereo version of ORB-SLAM \cite{mur2017orb} and \ov2slam on the EuRoC dataset with real-time enforced.  We show the trajectories obtained on the \textit{Machine Hall} (MHXX) sequences in Figure \ref{fig:euroc-stereo-mh} and on the \textit{Vicon Room} (VX-XX) sequences in Figure \ref{fig:euroc-stereo-vx}.

\begin{figure}[!ht]
    \begin{subfigure}[b]{0.475\linewidth}
        \centering \includegraphics[width=0.975\linewidth]{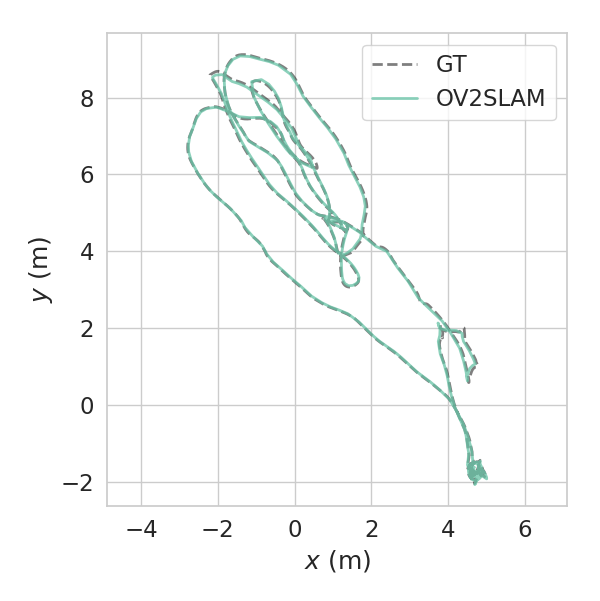}
        \captionsetup{justification=centering}
        \caption{EuRoC MH01}\label{fig:colmap_1}
    \end{subfigure}
    \begin{subfigure}[b]{0.475\linewidth}
        \centering \includegraphics[width=0.975\linewidth]{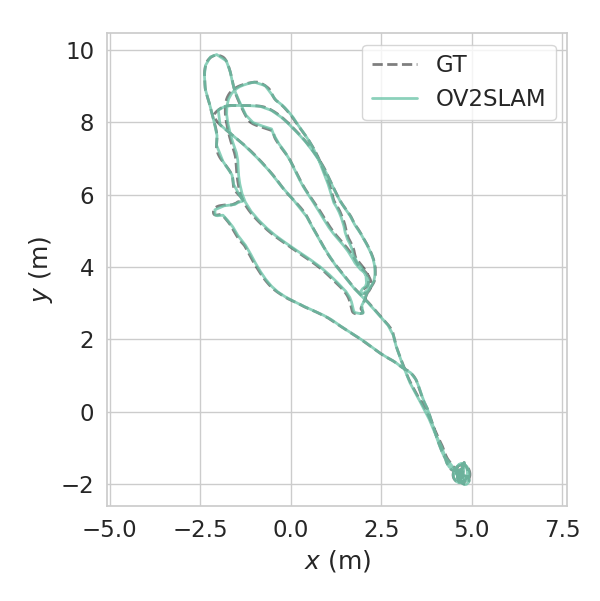}
        \captionsetup{justification=centering}
        \caption{EuRoC MH02}\label{fig:colmap_1}
    \end{subfigure}
    
    \begin{subfigure}[b]{0.475\linewidth}
        \centering \includegraphics[width=0.975\linewidth]{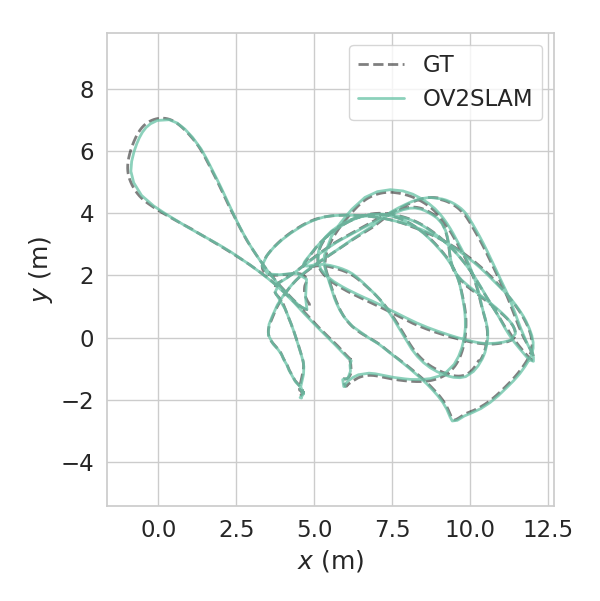}
        \captionsetup{justification=centering}
        \caption{EuRoC MH03}\label{fig:colmap_1}
    \end{subfigure}
    \begin{subfigure}[b]{0.475\linewidth}
        \centering \includegraphics[width=0.975\linewidth]{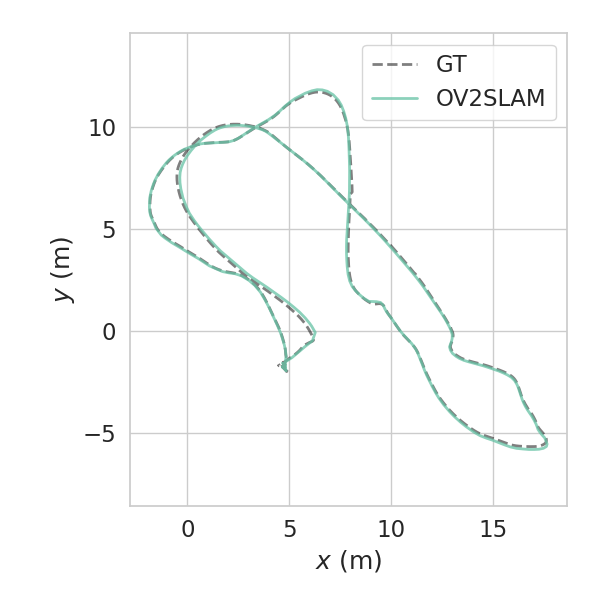}
        \captionsetup{justification=centering}
        \caption{EuRoC MH04}\label{fig:colmap_1}
    \end{subfigure}
    
    \begin{subfigure}[b]{0.475\linewidth}
        \centering \includegraphics[width=0.975\linewidth]{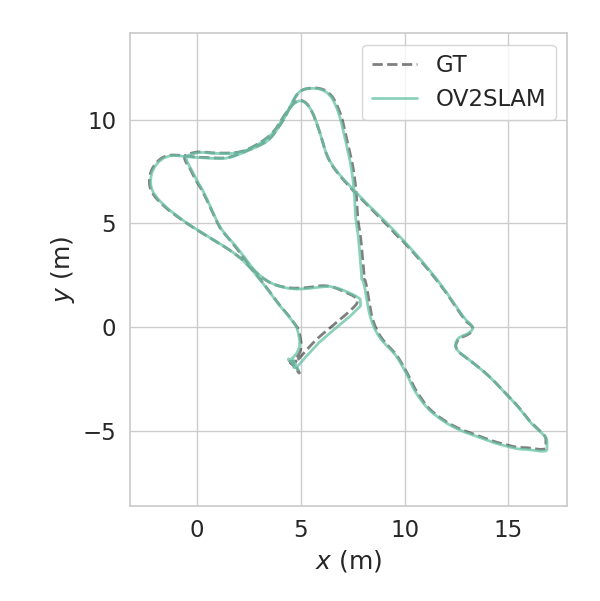}
        \captionsetup{justification=centering}
        \caption{EuRoC MH05}\label{fig:colmap_1}
    \end{subfigure}
    
    \captionsetup{justification=centering}
    \caption{Trajectories estimated in Real-Time with monocular \ov2slam without LC on EuRoC Machine Hall (MHXX) sequences.}
    
    \label{fig:euroc-mono-mh}
\end{figure}

\begin{figure}[!ht]
    \begin{subfigure}[b]{0.475\linewidth}
        \centering \includegraphics[width=0.975\linewidth]{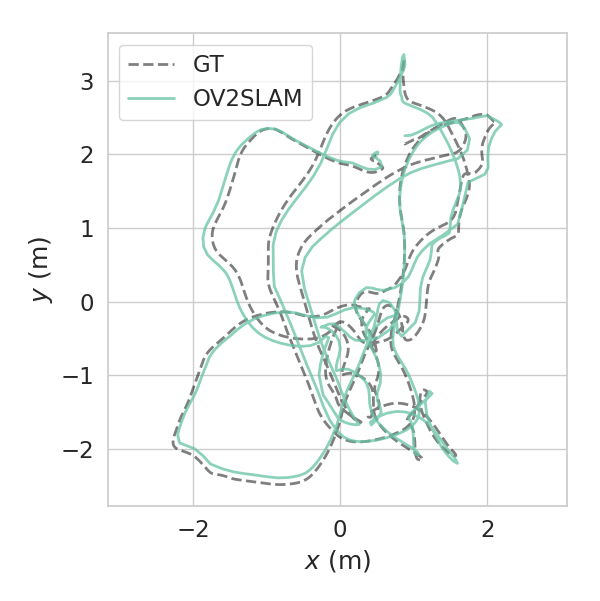}
        \captionsetup{justification=centering}
        \caption{EuRoC V1 01}\label{fig:colmap_1}
    \end{subfigure}
    \begin{subfigure}[b]{0.475\linewidth}
        \centering \includegraphics[width=0.975\linewidth]{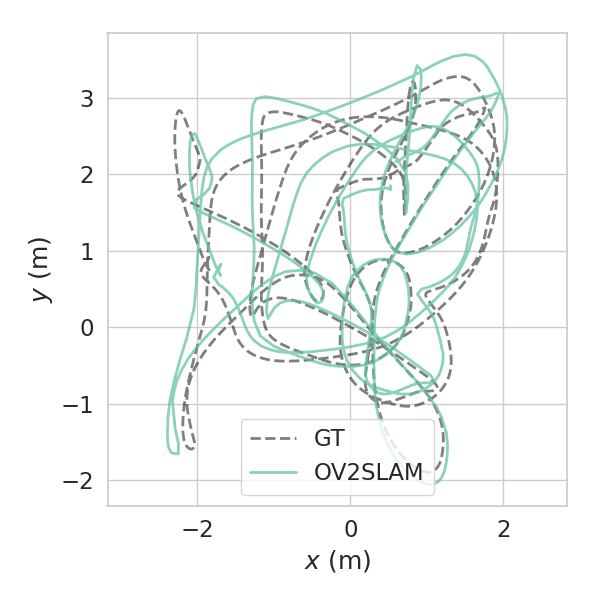}
        \captionsetup{justification=centering}
        \caption{EuRoC V1 02}\label{fig:colmap_1}
    \end{subfigure}
    
    \begin{subfigure}[b]{0.475\linewidth}
        \centering \includegraphics[width=0.975\linewidth]{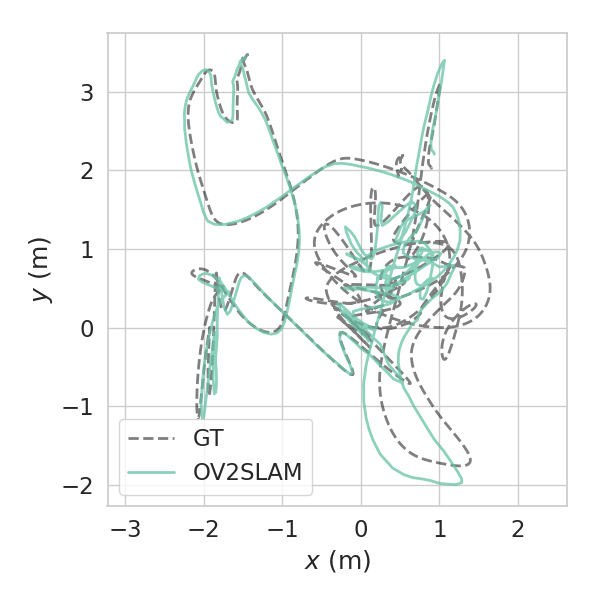}
        \captionsetup{justification=centering}
        \caption{EuRoC V1 03}\label{fig:colmap_1}
    \end{subfigure}
    \begin{subfigure}[b]{0.475\linewidth}
        \centering \includegraphics[width=0.975\linewidth]{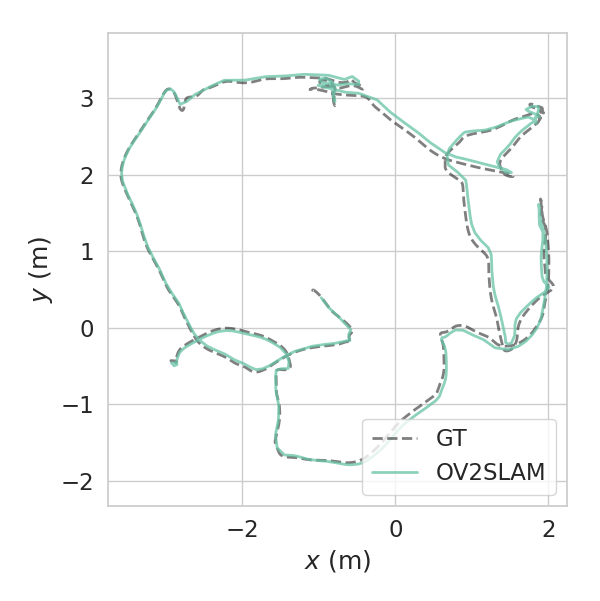}
        \captionsetup{justification=centering}
        \caption{EuRoC V2 01}\label{fig:colmap_1}
    \end{subfigure}

    \captionsetup{justification=centering}
    \caption{Trajectories estimated in Real-Time with monocular \ov2slam without LC on EuRoC Vicon Room (VX-XX) sequences.}
    
    \label{fig:euroc-mono-vx}
\end{figure}

We further display the trajectories estimated with the monocular version of \ov2slam with real-time processing enforced in Figure \ref{fig:euroc-mono-mh} and Figure \ref{fig:euroc-mono-vx}.

\section{KITTI Experiments}

\begin{figure*}
    \centering
    \begin{subfigure}[b]{0.3\linewidth}
        \centering \includegraphics[width=0.975\linewidth]{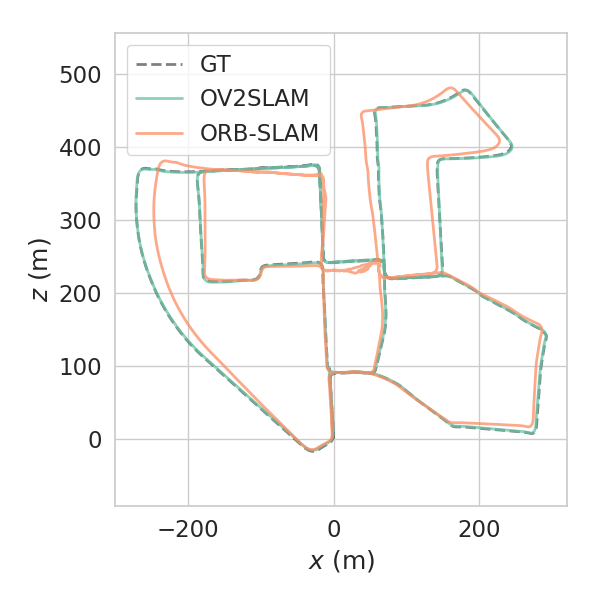}
        \captionsetup{justification=centering}
        \caption{KITTI 00}\label{fig:colmap_1}
    \end{subfigure}
    \begin{subfigure}[b]{0.3\linewidth}
        \centering \includegraphics[width=0.975\linewidth]{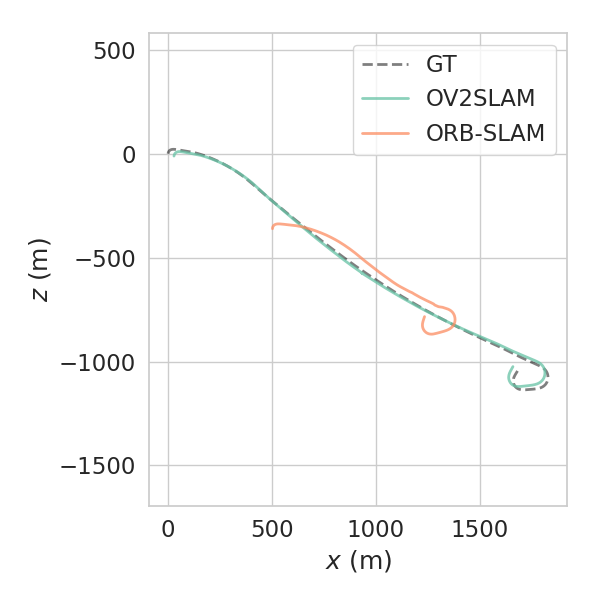}
        \captionsetup{justification=centering}
        \caption{KITTI 01}\label{fig:colmap_1}
    \end{subfigure}
    \begin{subfigure}[b]{0.3\linewidth}
        \centering \includegraphics[width=0.975\linewidth]{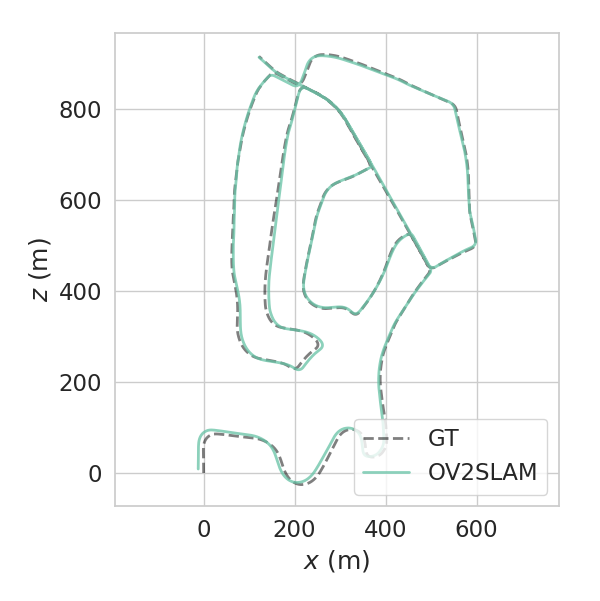}
        \captionsetup{justification=centering}
        \caption{KITTI 02}\label{fig:colmap_1}
    \end{subfigure}
    
    \begin{subfigure}[b]{0.3\linewidth}
        \centering \includegraphics[width=0.975\linewidth]{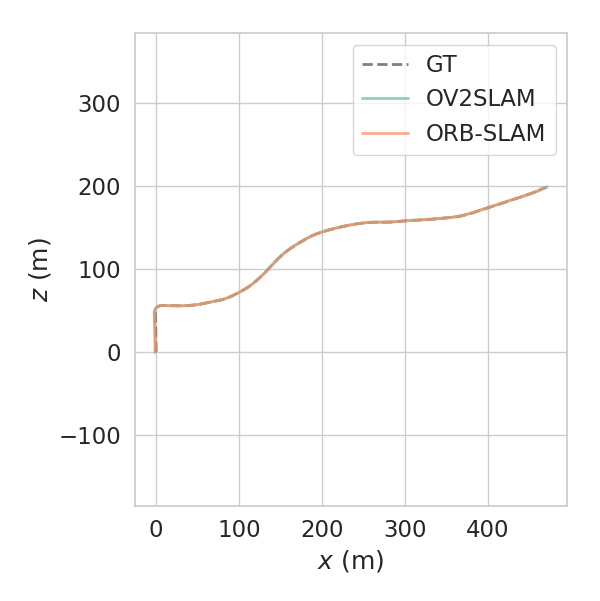}
        \captionsetup{justification=centering}
        \caption{KITTI 03}\label{fig:colmap_1}
    \end{subfigure}
    \begin{subfigure}[b]{0.3\linewidth}
        \centering \includegraphics[width=0.975\linewidth]{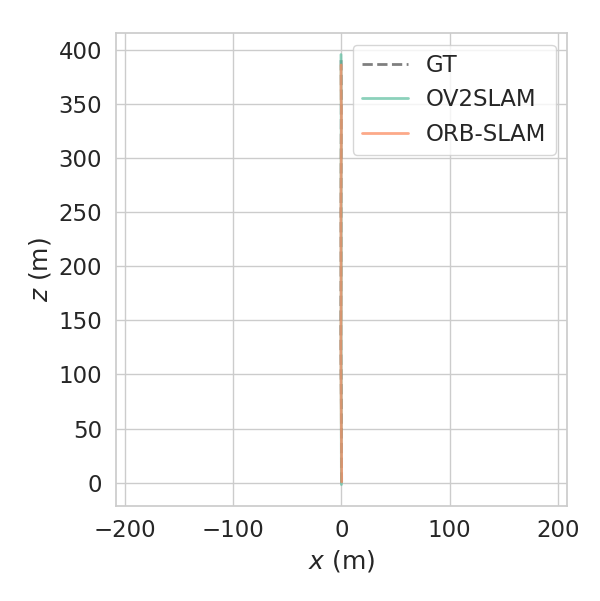}
        \captionsetup{justification=centering}
        \caption{KITTI 04}\label{fig:colmap_1}
    \end{subfigure}
    \begin{subfigure}[b]{0.3\linewidth}
        \centering \includegraphics[width=0.975\linewidth]{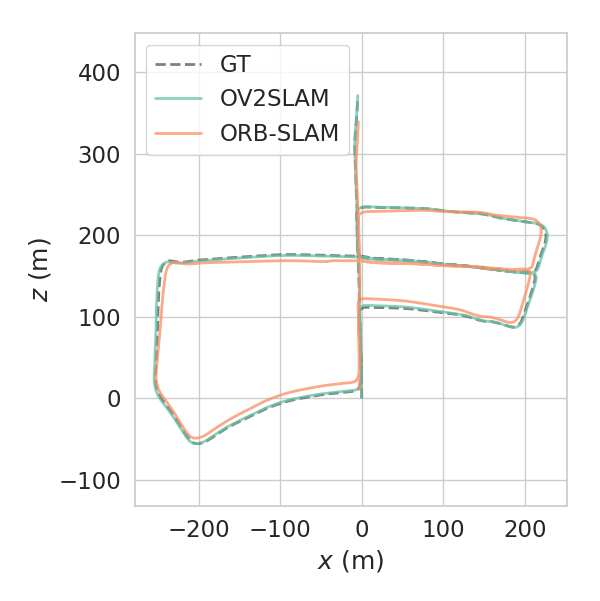}
        \captionsetup{justification=centering}
        \caption{KITTI 05}\label{fig:colmap_1}
    \end{subfigure}
    
    \begin{subfigure}[b]{0.3\linewidth}
        \centering \includegraphics[width=0.975\linewidth]{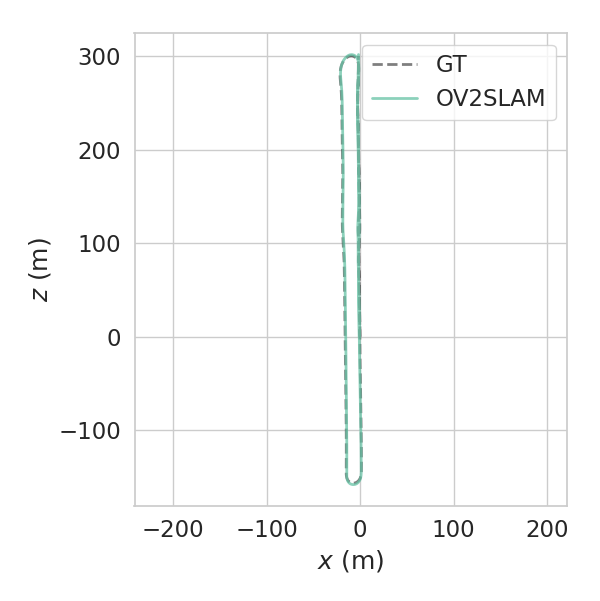}
        \captionsetup{justification=centering}
        \caption{KITTI 06}\label{fig:colmap_1}
    \end{subfigure}
    \begin{subfigure}[b]{0.3\linewidth}
        \centering \includegraphics[width=0.975\linewidth]{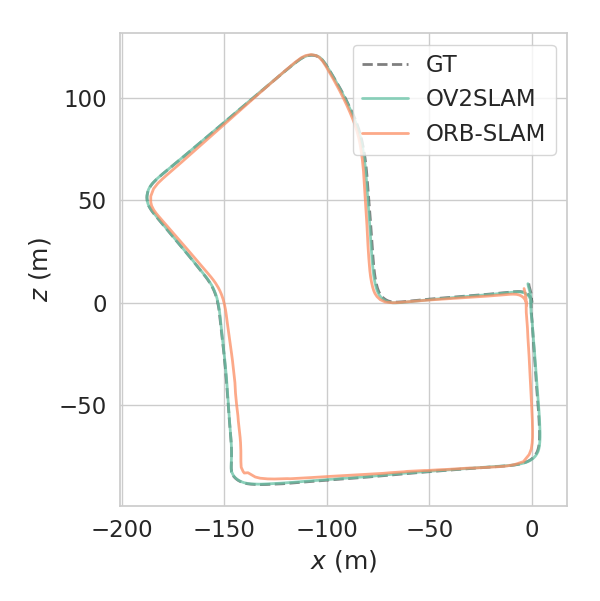}
        \captionsetup{justification=centering}
        \caption{KITTI 07}\label{fig:colmap_1}
    \end{subfigure}
    \begin{subfigure}[b]{0.3\linewidth}
        \centering \includegraphics[width=0.975\linewidth]{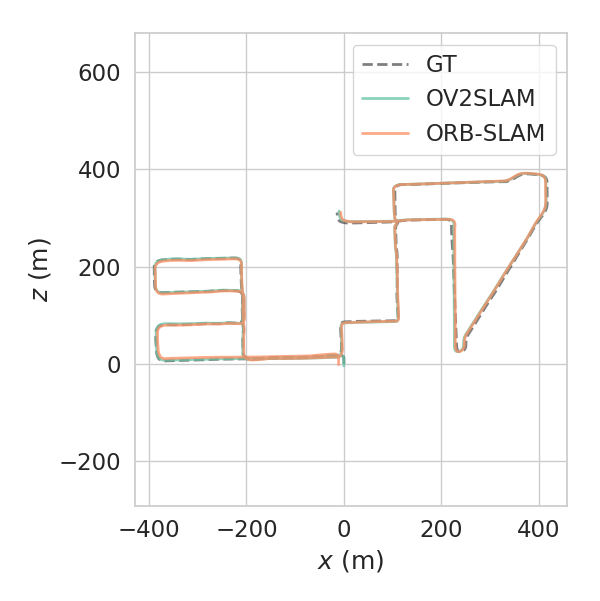}
        \captionsetup{justification=centering}
        \caption{KITTI 08}\label{fig:colmap_1}
    \end{subfigure}

    \begin{subfigure}[b]{0.3\linewidth}
        \centering \includegraphics[width=0.975\linewidth]{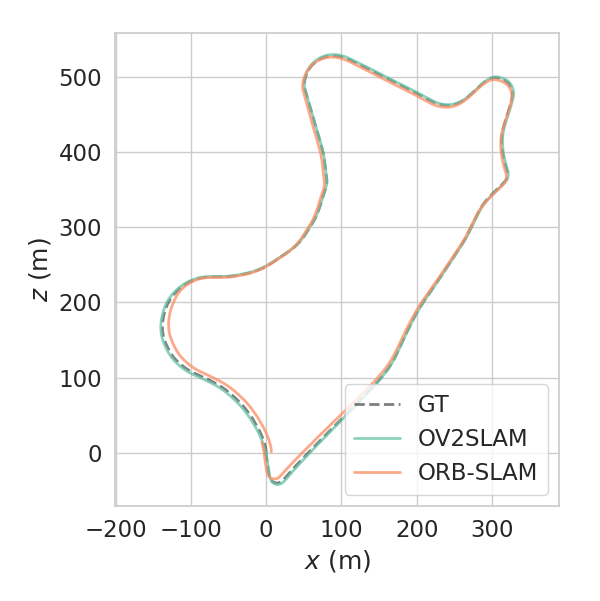}
        \captionsetup{justification=centering}
        \caption{KITTI 09}\label{fig:colmap_1}
    \end{subfigure}
    \begin{subfigure}[b]{0.3\linewidth}
        \centering \includegraphics[width=0.975\linewidth]{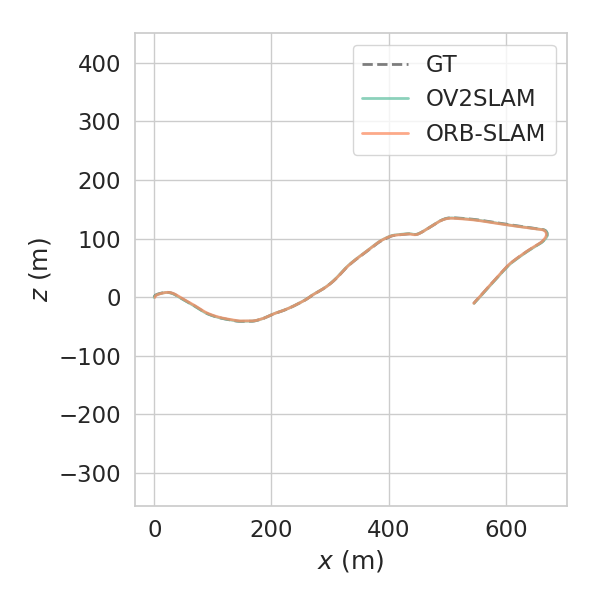}
        \captionsetup{justification=centering}
        \caption{KITTI 10}\label{fig:colmap_1}
    \end{subfigure}
    \begin{subfigure}[b]{0.3\linewidth}
    \end{subfigure}

    \captionsetup{justification=centering}
    \caption{Trajectories estimated in Real-Time with stereo \ov2slam and ORB-SLAM on KITTI training set.}
    \label{fig:kitti_trajs}
    
\end{figure*}

We display the trajectories obtained with both ORB-SLAM and \ov2slam while enforcing real-time on the KITTI dataset in Figure \ref{fig:kitti_trajs}.

\end{document}